\documentclass{article}

\usepackage{PRIMEarxiv}

\usepackage[utf8]{inputenc} 
\usepackage[T1]{fontenc}    
\usepackage{hyperref}       
\usepackage{url}            
\usepackage{booktabs}       
\usepackage{amsfonts}       
\usepackage{nicefrac}       
\usepackage{microtype}      
\usepackage{lipsum}
\usepackage{fancyhdr}       
\usepackage{graphicx}       
\graphicspath{{media/}}     

\title{Understanding Place Identity with Generative AI
}

\author{
  Kee Moon Jang \\
  MIT Senseable City Lab \\
  \texttt{keejang@mit.edu} \\
   \And
  Junda Chen \\
  DataChat \\
  \texttt{jchen693@wisc.edu} \\
  \AND
  Yuhao Kang\thanks{Corresponding author: yuhaokang@mit.edu} \\
  MIT Senseable City Lab \\
  \texttt{yuhaokang@mit.edu} \\
  \And
  Junghwan Kim \\
  Virginia Tech \\
  \texttt{junghwankim@vt.edu} \\
  \And
  Jinhyung Lee\\
  Western University \\
  \texttt{jinhyung.lee@uwo.ca} \\
  \And
  Fábio Duarte\\
  MIT Senseable City Lab \\
  \texttt{fduarte@mit.edu} \\
}

\begin{document}
\maketitle

\begin{abstract}
Researchers are constantly leveraging new forms of data with the goal of understanding  how people perceive the built environment and build the collective place identity of cities. Latest advancements in generative artificial intelligence (AI) models have enabled the production of realistic representations learned from vast amounts of data. In this study, we aim to test the potential of generative AI as the source of textual and visual information in capturing the place identity of cities assessed by filtered descriptions and images. We asked questions on the place identity of a set of 31 global cities to two generative AI models, ChatGPT and DALL·E2. Since generative AI has raised ethical concerns regarding its trustworthiness, we performed cross-validation to examine whether the results show similar patterns to real urban settings. In particular, we compared the outputs with Wikipedia data for text and images searched from Google for image. Our results indicate that generative AI models have the potential to capture the collective image of cities that can make them distinguishable. This study is among the first attempts to explore the capabilities of generative AI in understanding human perceptions of the built environment. It contributes to urban design literature by discussing future research opportunities and potential limitations.
\end{abstract}


\section{Introduction}
Place identity, often referred to as properties that distinguish a place from others \cite{relph1976place,proshansky2014place}, is an important concept in the fields of urban design, geography, tourism, and environmental psychology. As a sense of place is shaped through diverse human experiences, recognizing such place characteristics has been crucial for understanding human-environment interactions. There is a positive impact of understanding place identity in facilitating planning processes to create livable and legible places \cite{jang2019crowd,manzo2006finding,meenar2022analyzing}.
Yet, measuring and representing place identity has been a challenging task due to the intrinsically subjective nature of place identity. Conventional studies attempted to capture built environment characteristics and human perceptions through direct observation, questionnaires, surveys and interviews \cite{jang2019crowd,meenar2022analyzing}. In the past decade, researchers have been leveraging new data sources to understand the collective place identity of cities. In particular, two data formats, texts and images, have been employed to advance our knowledge of place identity. Natural language processing (NLP) methods have been utilized for tasks such as sentiment analysis and topic modeling, which enable us to understand individuals’ opinions and emotions of places from online text corpuses \cite{jang2019crowd,gao2017data,hu2019semantic}. Meanwhile, advanced machine learning and computer vision techniques have been effective in revealing visual information about places from street-level images and geotagged photos \cite{zhang2018representing,zhang2019discovering,zhang2020uncovering}. Progress in these approaches helps engage textual and image data at a larger scale for place-informative purposes. Urban planners and designers have benefited from these emerging data sources to explore subjective urban experiences and promote data-driven decision-making processes in practices \cite{meenar2022analyzing}.
Recently, advancements in generative artificial intelligence (GenAI) models have received significant attention due to their capabilities to generate realistic text and image output based on natural language prompts. ChatGPT and DALL·E2, for instance, have been highlighted as powerful tools with potential for a wide variety of applications in different domains such as education, transportation, geography, and so forth \cite{kim2023does,latif2023artificial,mai2023opportunities,van2023chatgpt}. Also, there have been attempts in urban studies to evaluate design qualities of the built environment scenes and obtain optimal land-use configuration through automated urban planning process \cite{seneviratne2022dalle,wang2023towards}. Despite its promise in urban science, the use of generative AI also faces common ethical concerns such as misinformation and bias, falling short in depicting composition and locales for specific conditions \cite{kang2023ethics}. Therefore, there remains need for more robust quantitative examination and analysis on how well they represent place-specific contexts toward trustworthy outputs in different domains.
To this end, since generative AI models are offering new ways to collect textual and visual information that may represent realistic human responses, this study aims to examine the potential of generative AI as new tools for understanding place identity in different cities. In this endeavor, we address two research questions: (1) Can generative AI models identify place identity of cities? and (2) How reliable are the generated outputs when compared with real-world settings? This study is expected to guide urban researchers in using such tools to generate large volumes of data through a more efficient and cost-effective approach, as well as to study place identity in a data-driven manner, which can facilitate our understanding of urban perception.

\begin{figure}[h]
    \centering
    \includegraphics[width=0.65\textwidth]{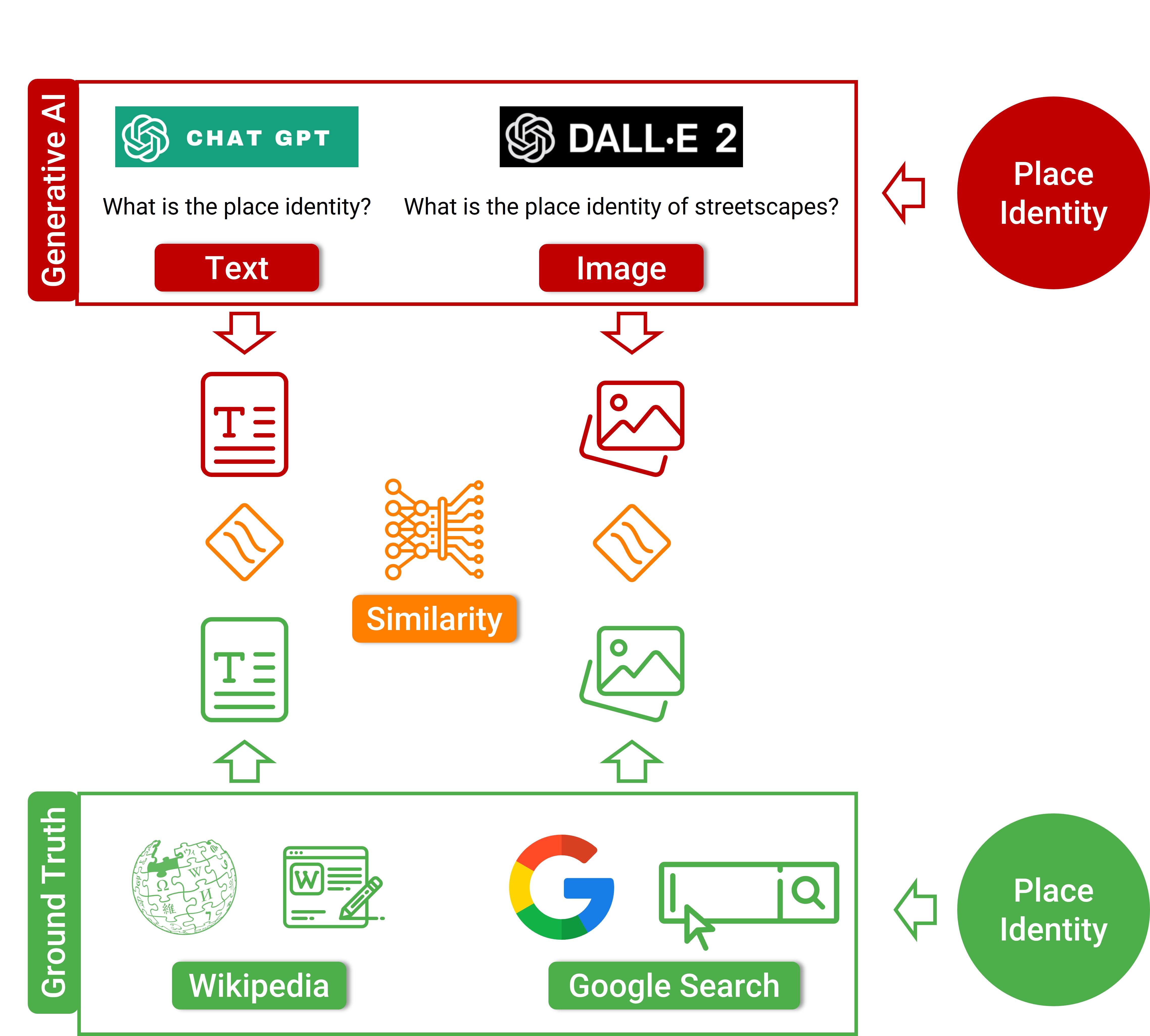}
    \caption{The computational framework of this paper}
    \label{fig:framework}
\end{figure}

\section{Methodology}

We present a computational framework of this study in Figure \ref{fig:framework}. The framework involves two primary datasets that we created to investigate the potential of generative AI models in capturing place identity of 31 global cities. The first dataset is a text-based dataset that we generated using ChatGPT to understand place identity, using a set of prompts in the following format: \textit{``What is the place identity of \{city\}? Give me in ten bullet points''.}
It should be noted that the place identity generated by ChatGPT may vary in length and style. To ensure consistency and comparability across different cities included in our dataset, we limited the responses to ten bullet points. By doing so, the generated outputs are concise and structured, and can easily be analyzed and compared.
The second dataset is an image-based dataset that we collected using DALL·E2 to generate visual representations of streetscapes in different cities. The prompts used to achieve this are the following:
\textit{``What is the place identity of streetscapes of \{city\}''?}
We generated 10 images for every city, where each image has a size of 256*256 pixels. By combining the image-based dataset with the text-based dataset, we aim to provide a comprehensive and multi-modal understanding of the place identity of each city.

We further collected two ground-truth datasets including a text dataset from Wikipedia and an image dataset from Google search. Despite the high performance in generative AI tools in generating realistic outputs, concerns regarding their reliability and accurate have emerged. Thus, we performed the cross-validation to compare the similarities among these datasets to evaluate whether results provided by generative AI can be trustworthy. For text similarity, we segmented the Wikipedia corpus into individual sentences using Natural Language Toolkit (NLTK) Python library and converted each sentence from both dataset into word embedding using a sentence transformer BERT model based on a modified version of MiniLM. Then, we measured cosine similarity for sentence embeddings from ChatGPT responses and Wikipedia corpuses to assess the relevance between the two datasets. We also created word cloud images of each city for a visual comparison between topics covered in ChatGPT and Wikipedia texts. For image similarity, we measured the Learned Perceptual Image Patch Similarity (LPIPS) \cite{zhang2018unreasonable} to assess the perceptual similarity of images generated by DALL·E2 and collected via Google search. The LPIPS metric evaluates the distance between different image patches and produces scores ranging from 0 to 1, where a lower score indicates greater similarity, and vice versa. Subsequently, we identify the top 3 similar Google images for each DALL·E2-generated image based on similarity scores. These analyses allow us to validate whether the results generated by generative AI models are consistent with the real-world urban settings of each city, providing valuable insights for urban design research and practice.

\section{Results}
\subsection{Results of place identity generated by ChatGPT}
\begin{figure}[h]
    \centering
    \includegraphics[width=0.75\textwidth]{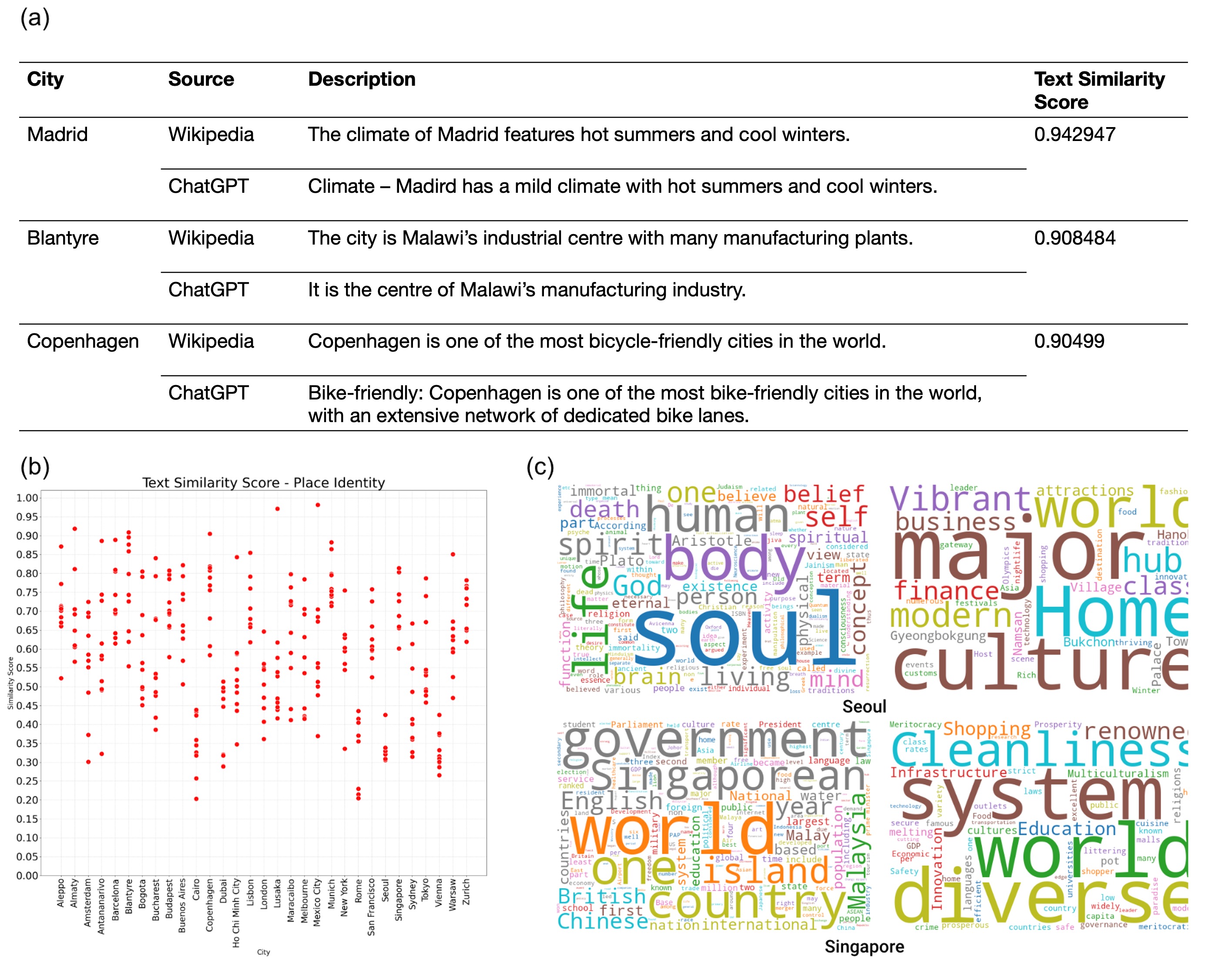}
    \caption{Text similarity results. (a) Examples of high text similarity scores between Wikipedia introductions and ChatGPT responses on place identity; (b) Scatter chart of the distribution of cosine similarity scores between sentences from ChatGPT responses and Wikipedia introduction corpuses; and (c) Word cloud comparison for Seoul and Singapore cases.}
    \label{fig:text_similarity}
\end{figure}

To validate the accuracy and reliability of the data generated by ChatGPT, we conducted cross-validation with Wikipedia. This involved computing the similarity scores between sentences from ChatGPT and Wikipedia, and presenting visual comparisons between pairs of word clouds. Figure \ref{fig:text_similarity} illustrates the validation results. Figure \ref{fig:text_similarity}(a) shows several examples of high sentence similarity scores. For example, for the city of Madrid, both Wikipedia and ChatGPT-generated sentences had similar descriptions of the climate, resulting in a very high similarity score of 0.94. However, as shown in Figure \ref{fig:text_similarity}(b), the comparison between ChatGPT-generated sentences and the introduction from Wikipedia resulted in a range of similarity scores, reflecting both similar and dissimilar descriptions of place identity. Such disparities may suggest that there are limitations to the effectiveness of generative AI models in capturing the nuances and complexities of place identity. Last, Figure \ref{fig:text_similarity}(c) illustrates two cases of word clouds analysis created for ChatGPT responses and Wikipedia. We found that ChatGPT captures the subjective atmosphere and cultural values as the most salient characteristics in Seoul, represented through topics including \textit{culture}, \textit{vibrant}, and \textit{modern}. These are intangible aspects of the capital city of South Korea, corresponding to the ‘meaning’ element of place identity models defined in fields of environmental psychology and geography \cite{canter1977psychology,relph1976place}. In the case of Singapore, we observe keywords such as \textit{government}, \textit{one country}, and \textit{system}. We interpret this as a representation of state governance in Singapore, implying that the text-to-text model identified its government structure.

\subsection{Results of place and urban identity generated by DALL·E2 }
\begin{figure}[h]
    \centering
    \includegraphics[width=0.75\textwidth]{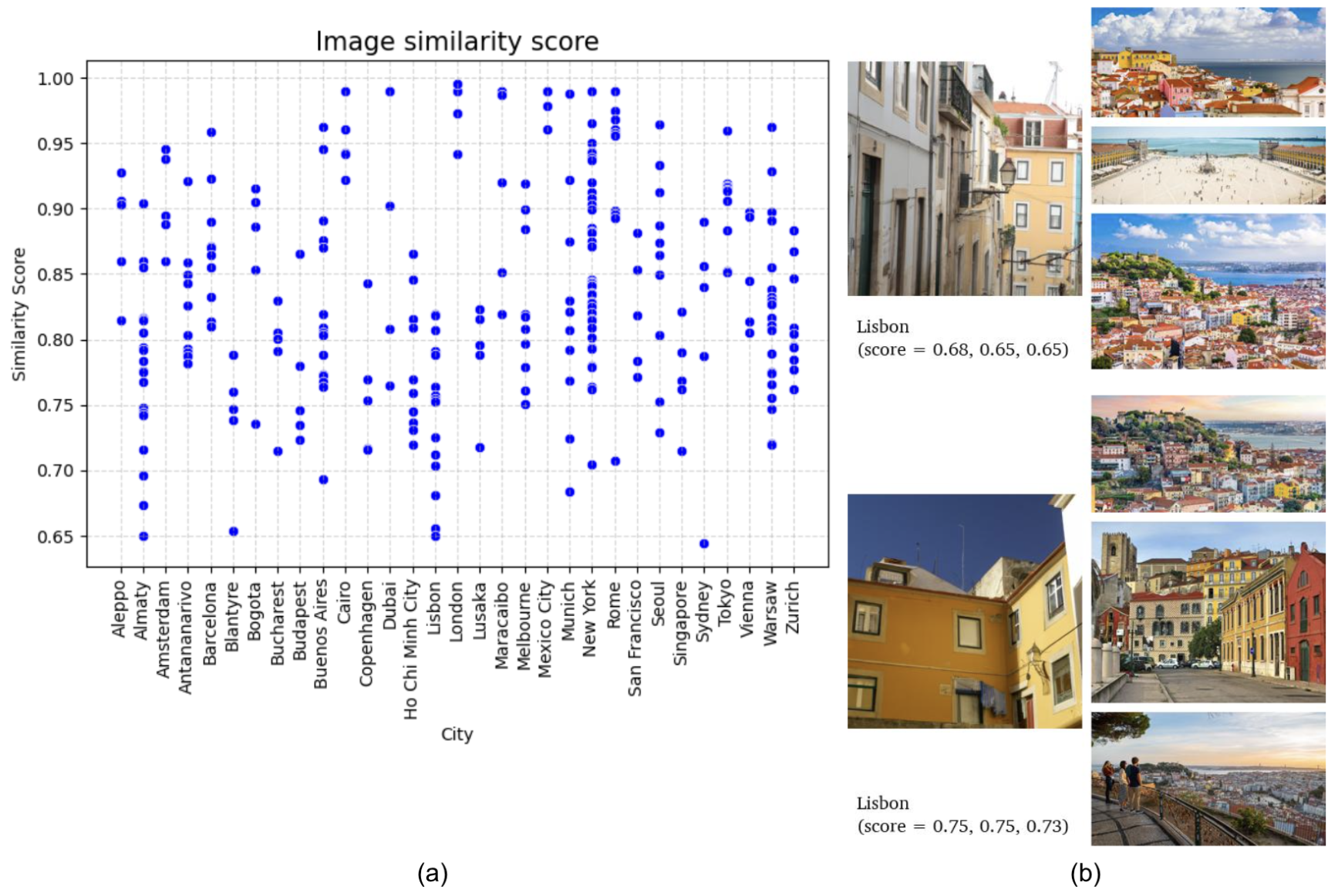}
    \caption{Image similarity results. (a) Scatter chart of the distribution of LPIPS scores between DALL·E2 generated images and Google image by cities; and (b) Low LPIPS score examples for Lisbon case.}
    \label{fig:img_similarity}
\end{figure}

Similar to the comparison between ChatGPT-generated sentences with Wikipedia corpus, we also compare images generated by DALL·E2 and those collected from Google image search. This was conducted to verify the reliability and generative capability of the text-to-image model in reproducing realistic representation of place-specific scenes of cities. For this purpose, image similarity score defined as LPIPS metric is calculated to assess the perceptual similarity between AI-generated and real-world images that matches well with human judgement. Figure \ref{fig:img_similarity} presents the image similarity results. Overall, as shown in Figure \ref{fig:img_similarity}(a), Almaty, Blantyre, Lisbon and Sydney were cities that reported the highest perceptual similarity with LPIPS value being approximately 0.65. In particular, Lisbon presents relatively consistent low similarity scores within the range of 0.65-0.82. Figure \ref{fig:img_similarity}(b) shows two examples of DALL·E2 generated images for Lisbon’s place identity and their top three matching Google image search results. It is evident that the generative AI effectively captured the low-rise residential buildings with vivid yellow colors in Lisbon, resulting in low LPIPS score (high similarity). These suggest that, despite variability across cities, DALL·E2 can generate more reliable images of urban scenes for certain cities that reflect their place identity.

\section{Conclusion}
In this study, we presented text and image similarity results between responses from two generative AI model, ChatGPT and DALL·E2, and corresponding ground-truth data to test the reliability of their outputs for representing place identity of different cities. Through examining the two datasets, we find that, in many cases, they generated text description or realistic images that represent salient characteristics of cities. In particular, text similarity scores aligned closely with similarities observed in sentence-by-sentence comparison and word clouds of ChatGPT responses and Wikipedia corpuses. This study is among the first to examine the capabilities of generative AI tools in representing the place identity of cities. The overall framework is expected to aid planners and designers in utilizing such tools to identify salient characteristics of cities for sustainable placemaking and city branding purposes. 
Despite the contributions of this study, we discuss potential limitations and research opportunities to be addressed in future studies. First, a portion of DALL·E2 generated images are still considered more generic than place-specific, which may not fully reflect the place identity. The outputs for different cities mostly depicted common urban features such as buildings, road signs, streetlights and pavements. These features are more relevant to the generic concept of a \textit{city}, rather than \textit{identity}, and fall short in representing the attributes that distinguish a particular city from the rest. Another limitation lies in the uncertainty in the image similarity results. While the LPIPS-based metric was used to evaluate the perceptual similarity between images in this study, we find cases whose interpretation remains obscure. In specific, similar scenes generated by DALL·E2 resulted in a range of similarity scores when measured against the same ground-truth image. Yet, it is uncertain why such differences are observed, what contributes to high or low similarity results, and thus which scene is most relevant to the place identity of a particular city.
Considering the limitations mentioned above, we outline future directions to be explored in further studies. Most importantly, careful prompt engineering can help obtain more reliable results that represent place-specific attributes of different cities. As suggestions to design effective prompts for DALL·E2 to yield more relevant responses, we can specify in regards to the point of view (POV), perspective, or captured objects of output images. Adding \textit{\{heading\}}, \textit{\{pitch\}}, \textit{\{perspective\}} or \textit{\{object\}} parameters as keywords in prompts can help obtain consistent viewpoints and minimize unpredictability in the scenes being generated by DALL·E2. Also, including keywords related to socioeconomic aspects (e.g., income level, race/ethnicity, etc.) enables discussing the fairness of generative AI models in representing the social context in urban settings. In the meantime, methods to evaluate the reliability of text and image outputs can be improved. A concrete threshold should be well defined to overcome the uncertainty in interpreting similarity scores. In addition, the generative model outputs can be validated against different ground-truth datasets. Social media texts and images are possible sources in this vein as they convey users’ various information related to places. Furthermore, different methods other than measuring similarity scores can be adopted for comparison purposes, such as detection algorithms to contrast object configurations derived from two datasets.


\bibliographystyle{unsrt}  
\bibliography{references}

\end{document}